\documentclass[letterpaper, 10 pt, conference]{ieeeconf}  

\usepackage{float} 
\usepackage{subfigure}
\usepackage{booktabs} 
\usepackage{amssymb}
\usepackage{cite}
\usepackage{amsmath}
\usepackage{threeparttable}
\usepackage{multirow}
\usepackage{graphicx}
\usepackage{bm}
\usepackage{multirow}
\usepackage{algorithm}
\usepackage{algpseudocode}
\usepackage{amsmath}

\IEEEoverridecommandlockouts                              

\title{\LARGE \bf
Sparse Optical Flow-Based Line Feature Tracking
}

\author{Qiang Fu$^{12}$, Hongshan Yu$^{1}$, Islam Ali$^{2}$, Hong Zhang$^{2}$,~\IEEEmembership{Fellow,~IEEE} 
\thanks{This work was supported in part by the National Natural Science Foundation of China under Grant 61973106 and Grant U1813205, U1913202, in part by the China Scholarship Council under Grant 201906130082, in part by the Key Research and Development Project of Science and Technology Plan of Hunan Province under Grant 2018GK2021, in part by the Key Project of Science and Technology Plan of Changsha City under Grant kq1801003. 
}
\thanks{$^{1}$The authors are with National Engineering Laboratory for Robot Visual Perception and Control Technology, Hunan University, China. Email: {\tt\small cn.fq@qq.com}}%
\thanks{$^{2}$The authors are with Department of Computing Science, University of Alberta, Canada. Email: {\tt\small hzhang@ualberta.ca}}%
}

\begin{document}

\maketitle
\thispagestyle{empty}
\pagestyle{empty}

\begin{abstract}
In this paper we propose a novel sparse optical flow (SOF)-based line feature tracking method for the camera pose estimation problem. This method is inspired by the point-based SOF algorithm and developed based on an observation that two adjacent images in time-varying image sequences satisfy brightness invariant. Based on this observation, we re-define the goal of line feature tracking: track two endpoints of a line feature instead of the entire line based on gray value matching instead of descriptor matching. To achieve this goal, an efficient two endpoint tracking (TET) method is presented: first, describe a given line feature with its two endpoints; next, track the two endpoints based on SOF to obtain two new tracked endpoints by minimizing a pixel-level grayscale residual function; finally, connect the two tracked endpoints to generate a new line feature. The correspondence is established between the given and the new line feature. Compared with current descriptor-based methods, our TET method needs not to compute descriptors and detect line features repeatedly. Naturally, it has an obvious advantage over computation. Experiments in several public benchmark datasets show our method yields highly competitive accuracy with an obvious advantage over speed. The source code of our method is released at: https://github.com/cnqiangfu/TET.
\end{abstract}

\section{Introduction}
Line-based camera pose estimation is gaining interest as the lines provide reliable constraints on the scene structure for high-accuracy demand, 
such as \cite{poseline, Li, pnl, lpcvpr, xupami}. And it is easy to be used to a visual odometer (VO) or simultaneous location and mapping (SLAM) system, such as \cite{fu2, yangtro, yangIROS, stuctvio, plvio, plvins, plslam, strifovio}. Therefore, it has a wide application range in robot location, navigation, map construction, and augmented reality \cite{zhang, vins, cnnsvo, fastorbslam, fu1}. \par

The performance of the line-based camera pose estimation methods relies on the accuracy and speed of line feature tracking \cite{linemgeo}. In this work, the tracking methods are focused, the porpose of which is to establish relable line feature correspondences. We observe that most of current methods track line features based on descriptors, which is developed based on a reasonable hypothesis that two line features, or, equivalently, image lines that observe the same space line should share the same descriptor. To be specific, these methods use the LSD \cite{lsd}, LBD\cite{lbd}, and KNN\cite{knn} algorithms to detect, describe and match line features. Naturally, the combination of LSD+LBD+KNN is recognized as a standard and state-of-the-art (SOTA) line feature tracking method. However, the descriptor-based methods are not efficient enough as they need to compute descriptors and detect line features for every frame, which has become a bottleneck for the real-time application due to their high computation demand \cite{yangIROS, plvio, yangtro}.  \par


\begin{figure}[t]
\centering  
	\subfigure[LSD+LBD+KNN]{
	\includegraphics[width=0.48\textwidth]{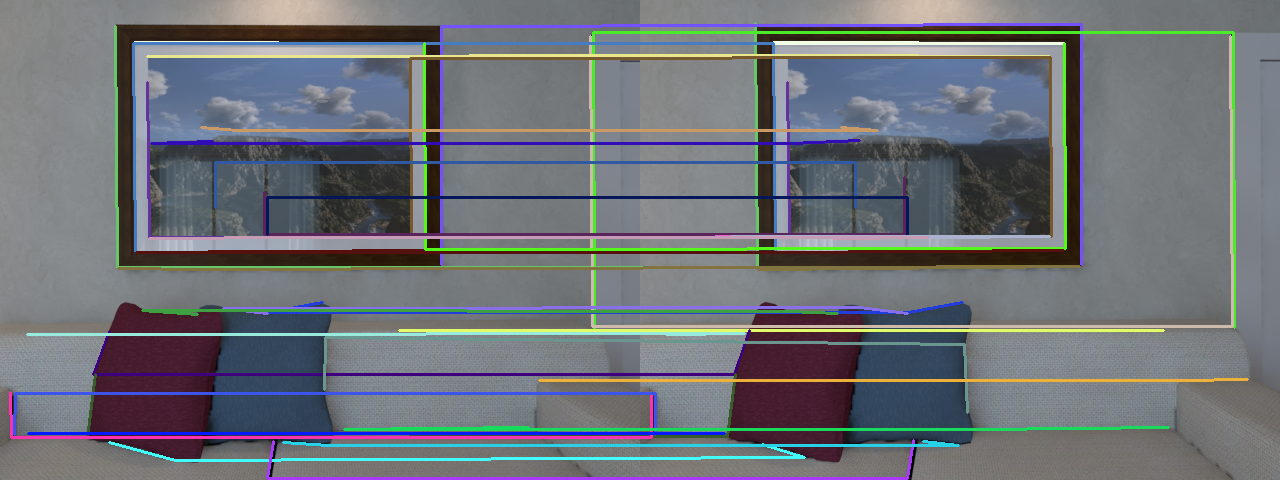}}\hspace{-1mm}
	\subfigure[LSD+TET (Ours)]{
	\includegraphics[width=0.48\textwidth]{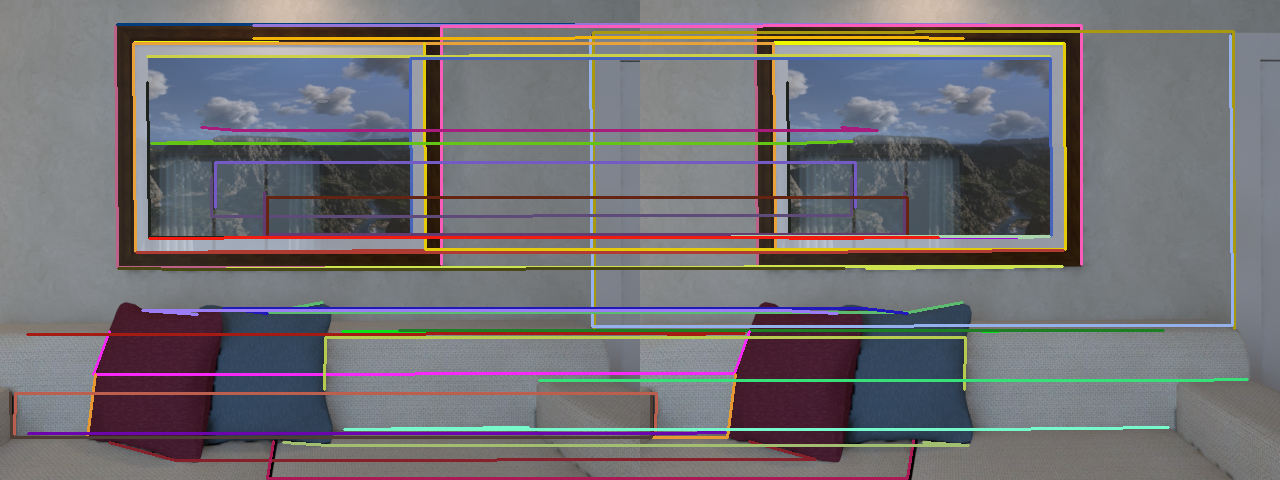}}	
	\quad
\caption{The comparsion of our method (LSD+TET) and LSD+LBD+KNN in two adjacent images which are from \textit{ICL-NUIM} dataset \cite{icl}. In the image of 640$\times$480 pixels, we discard short line features whose length is less than $30$ pixels as they are hard to match and unsuitable for the camera pose estimation problem \cite{plvins}. LSD detects $166$ line features with $43$ whose length is more than $30$ pixels. In the $43$ long line features, LBD+KNN tracks $36$ inliers while TET $42$. Our method finds more inliers.
}\label{f:figure1}
\end{figure}

\begin{table}[t]\caption{Time Static [ms] in Fig. \ref{f:figure1}}
\centering
\setlength{\tabcolsep}{2mm}{
\begin{tabular}{@{}l|ccccccc@{}}
\toprule
        & \multicolumn{2}{c|}{Left Image}  & \multicolumn{2}{c|}{Right Image} & \multicolumn{1}{c|}{\multirow{2}{*}{Match}} & \multirow{2}{*}{Total} \\
        & Det.  & \multicolumn{1}{c|}{Des.} & Det.  & \multicolumn{1}{c|}{Des.} & \multicolumn{1}{c|}{}                       &                        \\ \midrule
LSD+LBD+KNN & 14.42 & 12.95                     & 13.83 & 13.28                     & 1.63                                        & \textbf{55.03}\\
Ours    & 14.42 & --                        & --    & --                        & 2.37 & \textbf{16.79} \\ \bottomrule
\end{tabular}}
\begin{flushleft}
All statics are collected from real reproduction test. Det. represents detection, Des. represents description. Our method is much faster than the SOTA LSD+LBD+KNN method. 
\end{flushleft} 
\label{t:table1}  
\end{table}

Inspired by the fast point-based sparse optical flow (SOF) algorithm \cite{klt}, in this paper we present a novel light-weright SOF-based line feature tracking method which developed based on an observation that two adjacent frames in time-varying image sequence satisfy brightness invariant. Based on the observation, a hypothesis can be formulated that the corresponding points between two matched line features should share the same gray value. 
Further, by representing a line feature with its two endpoints, we re-define the goal of line feature tracking: tracking the two endpoints instead of all points of a line feature based on gray value matching instead of descriptor matching. 
\par

To achieve this goal, an efficient two endpoint tracking (TET) method is proposed: first, describe a given line feature with its two endpoints; next, track the two endpoints based on SOF to obtain two new tracked endpoints by minimizing a pixel-level grayscale residual function; finally, connect the two tracked endpoints to generate a new line feature. The correspondence is established between the given and the new line feature. Compared with dense optical flow, we only track the two endpoints rather than all points of a line feature. Compared with the descriptor-based methods, our TET method needs not to compute descriptors and detect line features repeatedly. As a result, it has an obvious advantage over computation, as Fig. \ref{f:figure1} shows.\par
Overall, our main contributions include:
\begin{itemize}
	\item To our best knowledge, this paper first defines the goal of line feature tracking as the two endpoints tracking. 	
	\item An efficient two endpoint tracking (TET) method is proposed for line feature tracking based on the SOF algorithm.
	\item Experiments in several public benchmark datasets show our method achieves the state-of-the-art (SOTA) accuracy with a prominent advantage over speed.
\end{itemize}
\par
In the remainder of the paper, we introduce related work in Section II, and the proposed TET method in Section III. The experiments are described in Section IV. Finally, the concluding remarks and future works are described in Section V.

\section{Related Work}
According to the matching algorithm, current line feature tracking methods are divided into descriptor-based and non-descriptor-based methods. In this section, we review the two groups of methods.

\subsection{Descriptor-Based Line Feature Tracking Methods}
\begin{figure}[h]
\centering  
	\subfigure[the descriptor-based method]{
	\includegraphics[width=0.48\textwidth]{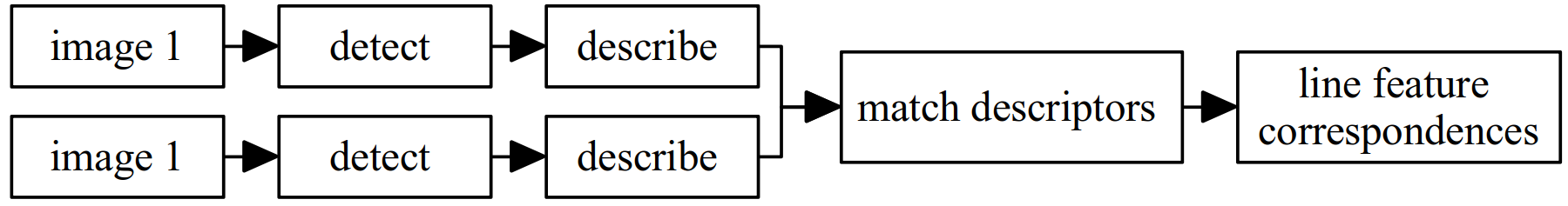}}
	\quad
	\subfigure[our gray-based method]{
	\includegraphics[width=0.48\textwidth]{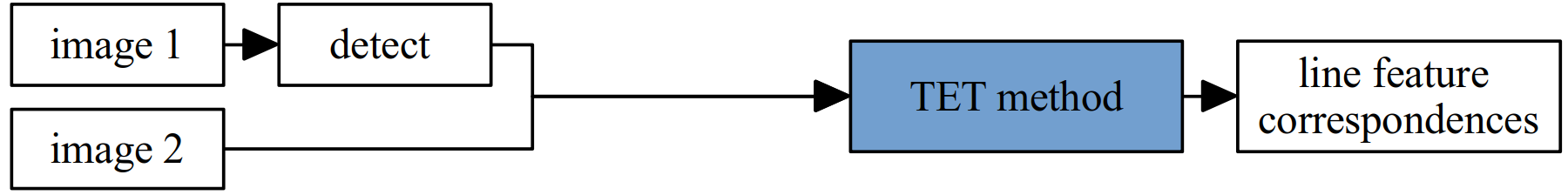}}		
\caption{A comparison of the standard descriptor-based method and our grayscale-based method. Note that our method needs not to compute descriptors or detect line features repeatedly.}
\label{f:methodcomparsion}
\end{figure}

The descriptor-based line feature tracking methods have attracted a great attention in the past several years because of their robustness even though with high computational cost, such as \cite{lsd, lbd, knn, detect1, detect2, detect3, detect4, tracking1, tracking2}. And they have been applied for the line-based camera pose estimation problem, such as \cite{fu2, yangtro, yangIROS, stuctvio, plvio, plvins, plslam, strifovio}. As Fig. \ref{f:methodcomparsion} shows, the process framework of this group of methods can be summarized into three steps: first, detect line features on the first (source) image and the second (target) image, respectively; second, compute the descriptors for all line features; finally, match these line features based on the descriptor distance.  
For the first step, detection, recent works tend to use deep learning technology to learn the line feature detection \cite{detect2, detect3, detect4} for scene representation. However, for the pose estimation problem, researchers still adopt the traditional hand-craft LSD algorithm \cite{lsd} for line detection as it needs less computation than the deep learning-based methods. For the second step and the third step, description and match, LBD\cite{lbd}+KNN\cite{knn} is widely used to describe and match for the pose estimation methods when incorporating line feature. Currently, the combination of LSD+LBD+KNN is considered as a standard and SOTA method in the domain of pose estimation \cite{poseline, Li, pnl, lpcvpr, xupami, fu2, yangtro, yangIROS, stuctvio, plvio, plvins, plslam, strifovio}.

\subsection{Non-Descriptor-Based Line Feature Tracking Methods}
Compared with the descriptor-based methods, the biggest of the Non-Descriptor-Based methods is not to compute descriptors for line features. One recent representative work is \cite{linemgeo}, in which Gomez-Ojeda \textit{et al.} proposed a pure-geometric-based algorithm for robust line feature tracking, which use a matching vector produced by the geometric constraint to evaluate similarity of two line features, obtaining SOTA performance in high dynamic range scenes. Note that this work also needs to detect line features in the target image. \par
In this work, we present a novel non-descriptor-based line feature tracking method which is inspired by the famous point-based SOP method \cite{klt} and developed based on a hypothesis that two line features that observe the same space line should share the same gray value, which we refer as gray-based method. This hypothesis is valid because two adjacent frames in a time-varying images sequence have two obvious features: short baseline and brightness invariant. Similar with SOP, our method also needs not to detect repeatedly and describe line features. Therefore, our method has a prominent advantage over computation.

\section{Method}
\label{s:method}

\begin{algorithm}[tp]
  \caption{Two Endpoint Tracking (TET) Method.}
  \label{ourmethod}
  \begin{algorithmic}[1]
    \Require
    The first frame $I_{1}$, the second frame $I_{2}$; 
    The line feature $L_{1}$ in $I_{1}$;          
    \Ensure
    New line feature $L_{2}$ in $I_{2}$ to establish line feature correspondence $L_{1}\leftrightarrow L_{2}$; 
    \State Extract two endpoints $L_{1}(x_{1},y_{1})$ and $L_{1}(x_{2},y_{2})$ of $L_{1}$;
    \State Represent $I_{1}$ and $I_{2}$ with $n$-level Gaussian image pyramid by Equation \ref{e:pyramid}, and describe $L_{1}(x_{1},y_{1})$ and $L_{1}(x_{2},y_{2})$ in the pyramid by Equation \ref{e:linepy};    
    \State Estimate movement vectors $\mathbf{m}_{1}$ and $\mathbf{m}_{2}$ of $L_{1}(x_{1},y_{1})$ and $L_{1}(x_{2},y_{2})$ in the pyramid, which is divided into three steps:
    \begin{enumerate}
	\item solve the movement vectors $\mathbf{m}_{1}^{n}$ and $\mathbf{m}_{2}^{n}$ in the deepest image $I_{2}^{n}$ (initial $\mathbf{m}_{1}^{n} = 0$, $\mathbf{m}_{2}^{n}=0$) via an extended iterative Lucas-Kanade method, for the iterative process:
	\begin{itemize}
	\item object function: Equation \ref{e:obejectfuction};
	\item gradient: Equation \ref{e:gradiant};
	\item termination condition: Equation \ref{e:termination}.
	\end{itemize}			
	\item propagate the computation results to $I_{2}^{n-1}$, and the results are regarded as two initial guesses for the movement vectors $\mathbf{m}_{1}^{n-1}$ and $\mathbf{m}_{2}^{n-1}$ in $I_{2}^{n-1}$;
	\item given the two initial guesses, refine the movement vectors in $I_{2}^{n-1}$, and propagate the results to $I_{2}^{n-2}$ and so on up to $I_{2}^{1}$ (original image), output $\mathbf{m}_{1}^{1}$ and $\mathbf{m}_{2}^{1}$;
	\end{enumerate}	
	\State Let $\mathbf{m}_{1}=\mathbf{m}_{1}^{1}= (dx_{1}^{},dy_{1}^{})$, $\mathbf{m}_{2}=\mathbf{m}_{2}^{1}= (dx_{2}^{},dy_{2}^{})$, and obtain the tracked two endpoints $L_{2}(x_{1}^{\prime}, y_{1}^{\prime}) = L_{2}(x_{1}+dx_{1}^{},y_{1}+dy_{1}^{})$ and $L_{2}(x_{2}^{\prime}, y_{2}^{\prime}) = L_{2}(x_{2}+dx_{2}^{},y_{2}+dy_{2}^{})$ in $I_{2}$; 
	\State Connect $L_{2}(x_{1}^{\prime}, y_{1}^{\prime})$ and $L_{2}(x_{2}^{\prime}, y_{2}^{\prime})$ to generate a new line feature $L_{2}$ in $I_{2}$;
	\\
    \Return $L_{2}$; (correspondence: $L_{1}\leftrightarrow L_{2}$)
  \end{algorithmic}  
\end{algorithm}

In this section we introduce the proposed sparse optical flow (SOF)-based line feature tracking method which is inspired by the famous point-based SOF algorithm \cite{klt}. One direct application of our method is the line-based VO or SLAM solutions. Different from the dense optical flow algorithm, our method only tracks the two endpoints rather than all points of a line feature. Different from the descriptor-based methods, our method needs not to compute descriptors and detect line features repeatedly. Same as SOF, our method relies on two hypotheses:\par
\begin{itemize}
	\item \textbf{Hypothesis 1}: gray invariant which means two matched line feature should share the same gray value. \par 
	\item \textbf{Hypothesis 2}: neighborhood motion consistency which means all points in the neighborhood of an endpoint keep the same movement as the endpoint. 
\end{itemize}
\par
The two hypotheses are valid because the consecutive frames in a vary-time image sequence has two features: short baseline and brightness invariant. Now, we introduce our method in detail.

\subsection{Goal State and Formulation}

\begin{figure}[tp]
\centering  
	\includegraphics[width=0.45\textwidth]{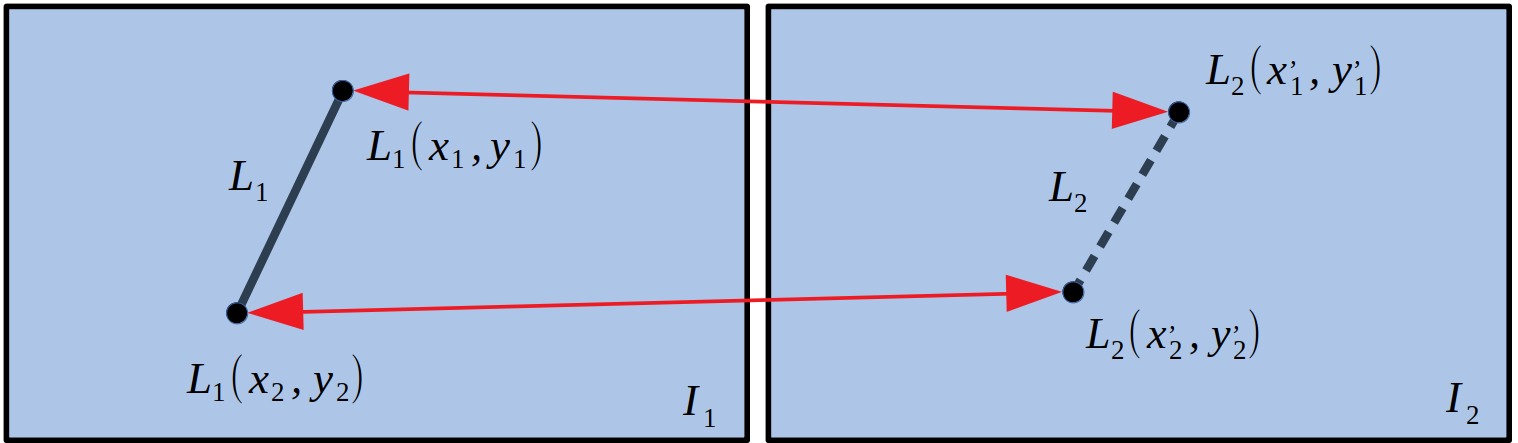}
\quad
\caption{The goal of line feature tracking. Given two images $I_{1}$ and $I_{2}$, and a line feature $L_{1}$ in $I_{1}$. Describe $L_{1}$ with its two endpoints $L_{1}(x_{1}^{}, y_{1}^{})$ and $L_{1}(x_{2}^{}, y_{2}^{})$. The goal of line feature tracking is to find two endpoints $L_{2}(x_{1}^{\prime}, y_{1}^{\prime})$ and $L_{2}(x_{2}^{\prime}, y_{2}^{\prime})$ in $I_{2}$. Aftet that, the correspondence will be established: $L_{1} \leftrightarrow L_{2}$.}
\label{f:linemodel}
\end{figure}

As Fig. \ref{f:linemodel} shows, by representing a line feature with its two endpoints, in this work we define the goal of line feature tracking as tracking the two points instead of all points of a line feature. In particular, the \textbf{goal} can be described as: \par
Given a line feature $L_{1}$ in the first (source) frame $I_{1}$, describe it with its two endponts $L_{1}(x_{1}^{}, y_{1}^{})$ and $L_{1}(x_{2}^{}, y_{2}^{})$. The tracking goal is to find two corresponding endpoints $L_{2}(x_{1}^{\prime}, y_{1}^{\prime})$ and $L_{2}(x_{2}^{\prime}, y_{2}^{\prime})$ in the second (target) frame $I_{2}$. Based on \textbf{Hypothesis 1}, we can formulate the goal: 


As Fig. \ref{linevector} shows, let $\mathbf{m}_{1}=(dx_{1}, dy_{1})$ be the movement vector from $L_{1}(x_{1}^{},y_{1}^{})$ to $L_{2}(x_{1}^{\prime},y_{1}^{\prime})$ and $\mathbf{m}_{2}=(dx_{2}, dy_{2})$ be the movement vector from $L_{1}(x_{2}^{},y_{2}^{})$ to $L_{2}(x_{2}^{\prime},y_{2}^{\prime})$. The goal can be converted to find two movement vectors $\mathbf{m}_{1}$ and $\mathbf{m}_{2}$ to simultaneously satisfy the two equations:
\begin{equation}
\begin{cases}I_{1}(L_{1}(x_{1}^{},y_{1}^{})) = I_{2}(L_{1}(x_{1}^{}+dx_{1}^{},y_{1}^{}+dy_{1}^{})) \\ 
I_{1}(L_{1}(x_{2}^{},y_{2}^{})) = I_{2}(L_{1}(x_{2}^{}+dx_{2}^{},y_{2}^{}+dy_{2}^{})) \end{cases},
\label{line1} 
\end{equation}
where $I(x,y)$ denote the gray value of $(x,y)$ on the image $I$. After that, we have $L_{2}(x_{1}^{\prime},y_{1}^{\prime}) = L_{1}(x_{1}^{}+dx_{1}^{},y_{1}^{}+dy_{1}^{})$ and $L_{2}(x_{2}^{\prime},y_{2}^{\prime}) = L_{1}(x_{2}^{}+dx_{2}^{},y_{2}^{}+dy_{2}^{})$. 
\par

\subsection{Two Endpoint Tracking}
\begin{figure}[tp]
\centering  
	\includegraphics[width=0.45\textwidth]{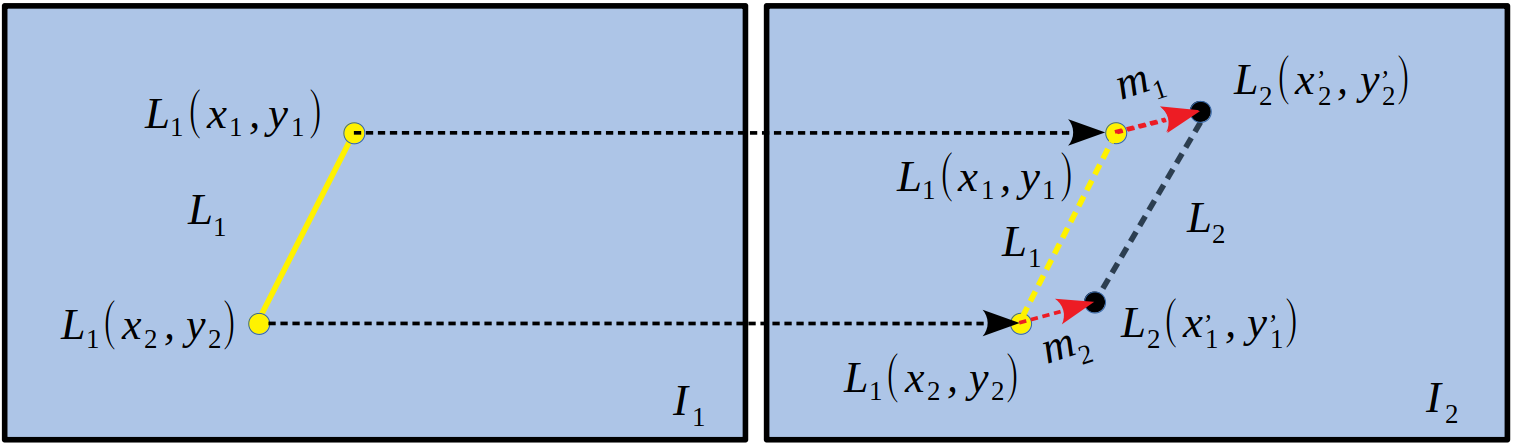}
\quad
\caption{Goal Formulation. Let $\mathbf{m}_{1}$ be the movement vector from $L_{1}(x_{1}^{},y_{1}^{})$ to $L_{2}(x_{1}^{\prime},y_{1}^{\prime})$, and $\mathbf{m}_{2}$ from $L_{1}(x_{2}^{},y_{2}^{})$ and $L_{2}(x_{2}^{\prime},y_{2}^{\prime})$. Based on \textbf{Hypothesis 1}, the goal is to find two movement vectors $\mathbf{m}_{1}$ and $\mathbf{m}_{2}$ to establish Equation (\ref{line1}) .
}
\label{linevector}
\end{figure}


To achieve the \textbf{goal}, we propose an efficient gray-based two endpoint tracking (TET) method, as \textbf{Algorithm} \ref{ourmethod} shows, which is divided into three steps: first, describe a given line feature with its two endpoints; next, track the two endpoints based on SOF to obtain two new tracked endpoints by minimizing a pixel-level gray residual function; finally, connect the two tracked endpoints to generate a new line feature. After that, the correspondence is established between the given and the new line feature. The key step is movement vector solving. Next, we will describe it.
\par


Let's take one endpoint as an example, we need to find a movement vector $\mathbf{m}_{1}=(dx_{1}, dy_{1})$ to satisfy equation:
\begin{equation}
I_{1}(L_{1}(x_{1}^{},y_{1}^{})) = I_{2}(L_{1}(x_{1}^{}+dx_{1}^{},y_{1}^{}+dy_{1}^{})),
\label{line2} 
\end{equation}
where $L_{1}(x_{1}^{}+dx_{1}^{},y_{1}^{}+dy_{1}^{}) = L_{2}(x_{1}^{\prime},y_{1}^{\prime})$; however, there is one problem that the above equation has two unknown variables $dx_{1}^{}$ and $dy_{1}^{}$. Therefore, we cannot solve it. For this problem, we need \textbf{Hypothesis 2} which indicates that all pixels in a $\omega$-size of window should keep consistent movement $\mathbf{m}_{1}$ as the center point $L_{1}(x_{1},y_{1})$ of the window. $\omega$ is an odd number and denotes the radius of the window. Thus we have:
\begin{equation}
\begin{cases}
\begin{aligned}
&I_{1}(L_{1}(x_{1}^{1},y_{1}^{1})) = I_{2}(L_{1}(x_{1}^{1}+dx_{1}^{},y_{1}^{1}+dy_{1}^{}))
\\ 
&I_{1}(L_{1}(x_{1}^{2},y_{1}^{2})) = I_{2}(L_{1}(x_{1}^{2}+dx_{1}^{},y_{1}^{2}+dy_{1}^{}))
\\
& \cdot\cdot\cdot\cdot\cdot\cdot\cdot\cdot\cdot\cdot\cdot\cdot\cdot\cdot\cdot\cdot\cdot\cdot\cdot\cdot\cdot\cdot\cdot\cdot
\\ 
& I_{1}(L_{1}(x_{1}^{k},y_{1}^{k})) = I_{2}(L_{1}(x_{1}^{k}+dx_{1}^{},y_{1}^{k}+dy_{1}^{}))
\end{aligned}
\end{cases},
\label{movementsolve} 
\end{equation}
where $k$ represents the number of all pixels in the $\omega$ -size of window, and the window center is $L_{1}(x_{1},y_{1})$. Note that Equation (\ref{movementsolve}) is over-determined. Therefore, the problem is now how to solve two unknown variables with $k>2$ equations, which can be solved via the method of least squares approximation. In this work, we solve it via a pixel-level gray residual error minimization method. Now we introduce the method model process. Notably, to simplify expressions, let $(x_{1},y_{1}) = L_{1}(x_{1},y_{1})$ and $(x_{2},y_{2}) = L_{1}(x_{2},y_{2})$, where we discard the $L$ sign. \par

\textbf{First}, we define the gray residual function. Let residual function of the two endpoints be $\epsilon(\mathbf{m}_{1})$ and $\epsilon(\mathbf{m}_{2})$, respectively. Based on Equation \ref{movementsolve}, we have:
\begin{equation}
\begin{cases}
\begin{aligned}
& \epsilon(\mathbf{m}_{1}) 
=\sum_{i=1}^k(I_{1}(x_{1}^{i},y_{1}^{i})-I_{2}(x_{1}^{i}+dx_{1}^{}, y_{1}^{i}+dy_{1}^{}))  \\
& \epsilon(\mathbf{m}_{2}) 
=\sum_{i=1}^k(I_{1}(x_{2}^{i},y_{2}^{i})-I_{2}(x_{2}^{i}+dx_{2}^{}, y_{2}^{i}+dy_{2}^{}))
\end{aligned}
\end{cases},
\end{equation}
where $i = [1, ..., k]$. \par
\textbf{Second}, the gray residual function is iteratively minimized to find the optimal movement vectors, which can be modeled as:
\begin{equation}
\begin{cases} \bm{m_{1}}(dx_{1}, dy_{1}) = \mathop{\arg\min}_{\bm{m_{1}}}  \left|\left|\epsilon(\bm{m_{1}})\right|\right|^{2} 
\\ \bm{m_{2}}(dx_{2}, dy_{2}) = \mathop{\arg\min}_{{\bm{m_{2}}}} \left|\left|\epsilon(\bm{m_{2}})\right|\right|^{2}
\end{cases}.
\label{e:obejectfuction}
\end{equation}
\par
For the above Equation \ref{e:obejectfuction}, in this work we use an extended iterative Lucas-Kanade method \cite{klt} to solve. The biggest difference is that the object of our method is two endpoints of a line feature while the object of the Lucas-Kanade method is one point.
The gradient expression of the solver process is computed by:
\begin{equation}
\begin{cases}
\frac{\partial \epsilon(\bm{m_{1}})}{\partial \bm{m_{1}}}=\sum_{i=1}^k\begin{bmatrix}I_{x}^{1}(x,y)I_{x}^{1}(x,y) & I_{x}^{1}(x,y) I_{y}^{1}(x,y)\\I_{x}^{1}(x,y) I_{y}^{1}(x,y) & I_{y}^{1}(x,y)I_{y}^{1}(x,y) \end{bmatrix}
\\\\ \frac{\partial \epsilon(\bm{m_{2}})}{\partial \bm{m_{2}}}=\sum_{i=1}^k\begin{bmatrix}I_{x}^{2}(x,y)I_{x}^{2}(x,y) & I_{x}^{2}(x,y) I_{y}^{2}(x,y)\\I_{x}^{2}(x,y) I_{y}^{2}(x,y) & I_{y}^{2}(x,y)I_{y}^{2}(x,y) \end{bmatrix}
\end{cases}
\label{e:gradiant}
\end{equation}
where
\begin{equation}
\begin{aligned}
&I_{x}^{1}(x,y) = \frac{\partial I_{1}(x,y)}{\partial x_{1}}= \frac{I_{1}(x_{1}^{i}+1,y_{1}^{})-I_{1}(x_{1}^{i}-1,y_{1}^{})}{2}
\\ & I_{y}^{1}(x,y) = \frac{\partial I_{1}(x,y)}{\partial y_{1}} = \frac{I_{1}(x_{1}^{},y_{1}^{i}+1)-I_{1}(x_{1}^{},y_{1}^{i}-1)}{2}
\\& I_{x}^{2}(x,y) = \frac{\partial I_{1}(x,y)}{\partial x_{2}}= \frac{I_{1}(x_{2}^{i}+1,y_{2}^{})-I_{1}(x_{2}^{i}-1,y_{2}^{})}{2}
\\& I_{y}^{2}(x,y) = \frac{\partial I_{1}(x,y)}{\partial y_{2}} = \frac{I_{1}(x_{2}^{},y_{2}^{i}+1)-I_{1}(x_{2}^{},y_{2}^{i}-1)}{2}
\end{aligned}
\end{equation}
where $I_{x}^{1}$ and $I_{y}^{1}$ denote the image derivatives at the position $L_{1}(x_{1},y_{1})$, $I_{x}^{2}$ and $I_{y}^{2}$ denote the image derivatives at position $L_{1}(x_{2},y_{2})$. The image derivatives $I_{x}$ and $I_{y}$ is computed directly from the first image $I_{1}$ in the neighborhood of the point independently from the second image $I_{2}$. In particular, we set $\omega = 7$, which means $k=49$. The iterative minimization process usually takes about $3-4$ ms to converge after 6 iterations, as Fig. \ref{f:iteratation} shows.

\par

\textbf{Finally}, when will the iterative minimization process terminate? In this work, we design an accuracy evaluation function to determine the termination condition. Let $\epsilon(\omega)$ denote the average gray residual function between two corresponding windows (patches). For two endpoints of a line feature, we have:
\begin{equation}
\begin{cases}
\begin{aligned}
&\epsilon(\omega_{1})=\frac{\sum_{i=1}^k{I_{1}(x_{1}^{k},y_{1}^{k})- I_{2}(x_{1}^{k}+dx_{1}^{},y_{1}^{k}+dy_{1}^{})}}{k}
\\& \epsilon(\omega_{2})= \frac{\sum_{i=1}^k{I_{1}(x_{2}^{k},y_{2}^{k}) - I_{2}(x_{2}^{k}+dx_{2}^{},y_{2}^{k}+dy_{2}^{})}}{k}
\end{aligned}.
\end{cases}
\end{equation} 
And then let $N_{iter}$ be number of iterations, the termination condition of the iterative optimization is modeled as:
 \begin{equation}
\left\{(\epsilon(\omega_{1}) \&  \epsilon(\omega_{2}) < \omega_{errormin}\right\} \| N_{iter} > N_{itermax}
\label{e:termination}
\end{equation}
where $\omega_{errormin}$ denotes the minimum value of the window error and $N_{itermax}$ denotes the maximum value of iterations. If $N_{iter} > N_{itermax}$ while $\epsilon(\omega_{1})$ or $\epsilon(\omega_{2}) > \omega_{errormin}$, we consider the line feature as an outlier. In this work, although the iterative process usually converges after 5 iterations, we set $\omega_{errormin} = 0.02$ and $N_{itermax} = 10$ in consideration of generality, as Fig. \ref{f:iteratation} shows.

\par

\begin{figure*}[t]
\centering  
	\subfigure[LSD+LBD+KNN]{
	\includegraphics[width=0.128\textheight]{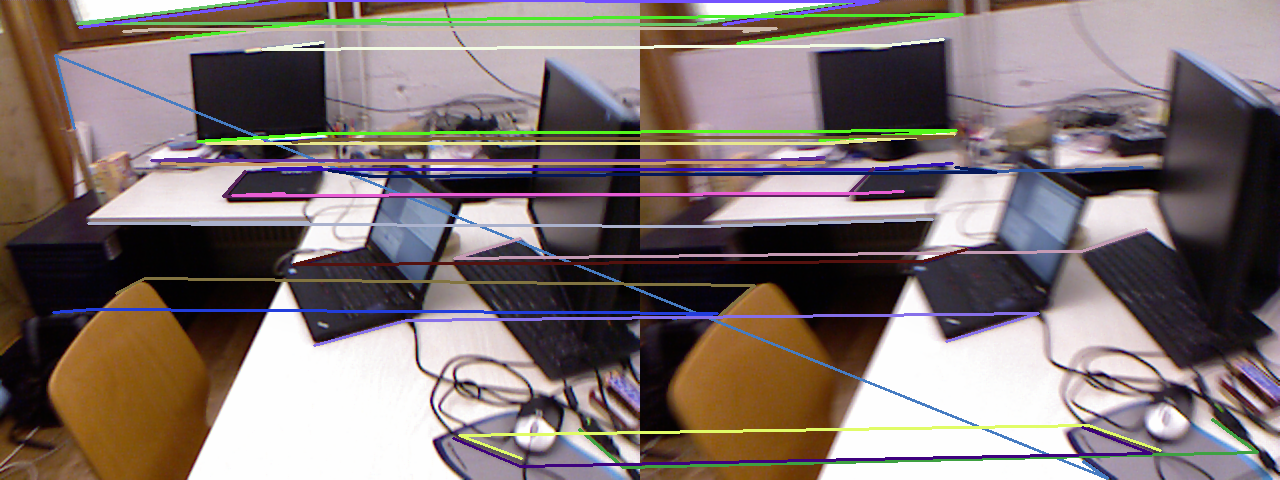}
	\includegraphics[width=0.128\textheight]{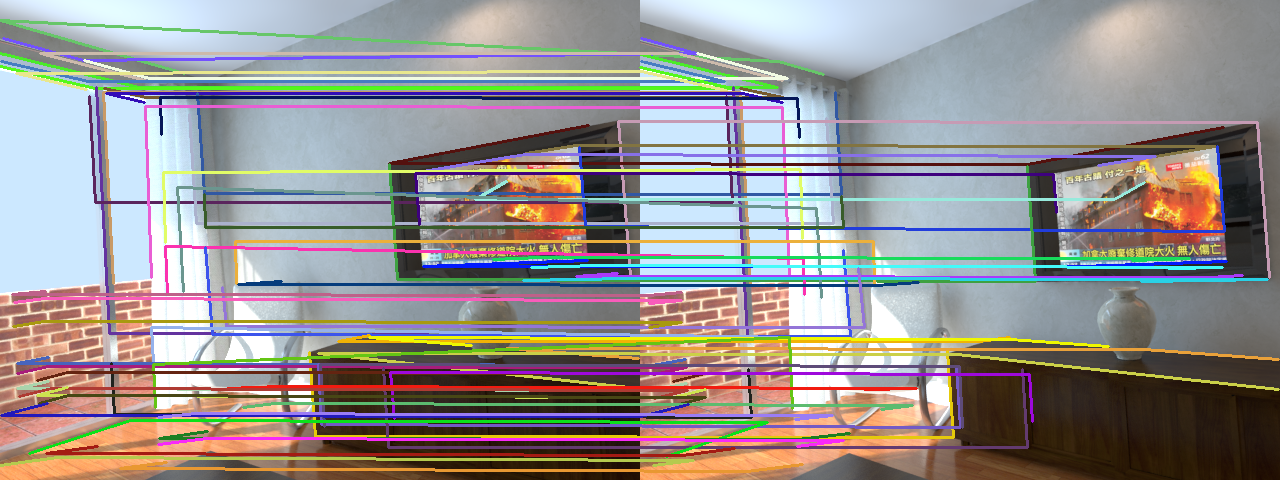}
	\includegraphics[width=0.150\textheight]{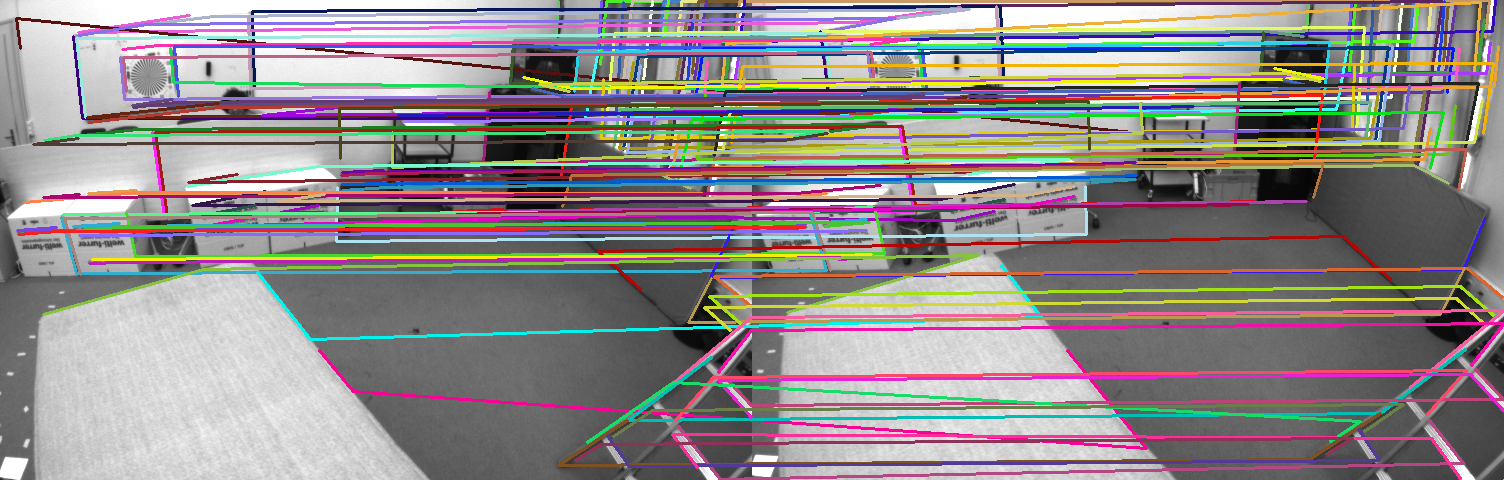}
	\includegraphics[width=0.315\textheight]{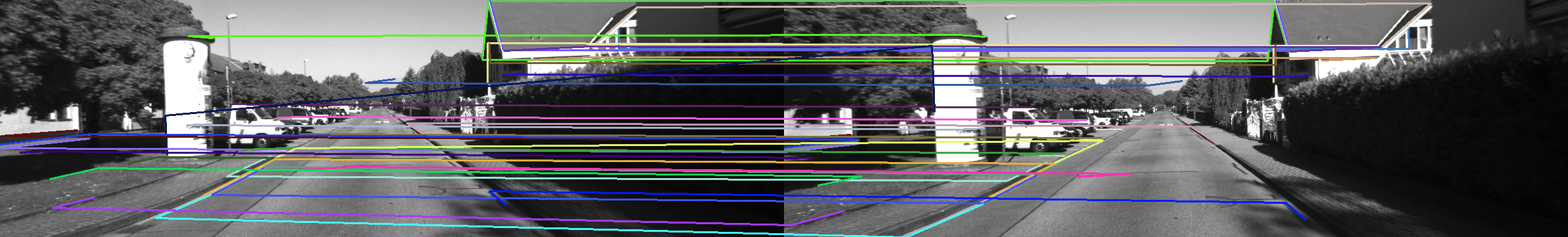}}
	
	\subfigure[LSD+TET (ours)]{
	\includegraphics[width=0.128\textheight]{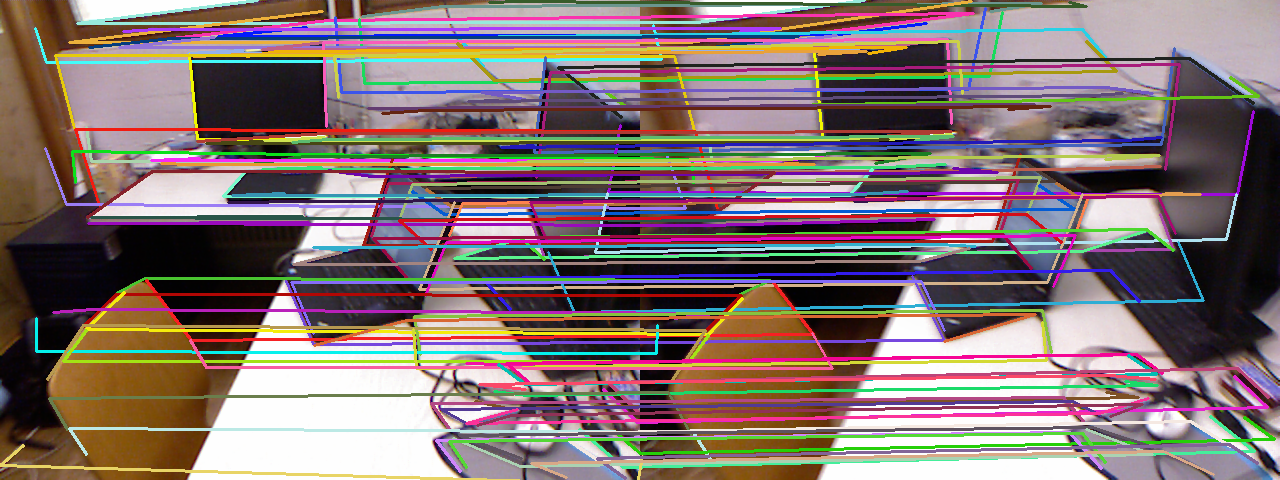}	
	\includegraphics[width=0.128\textheight]{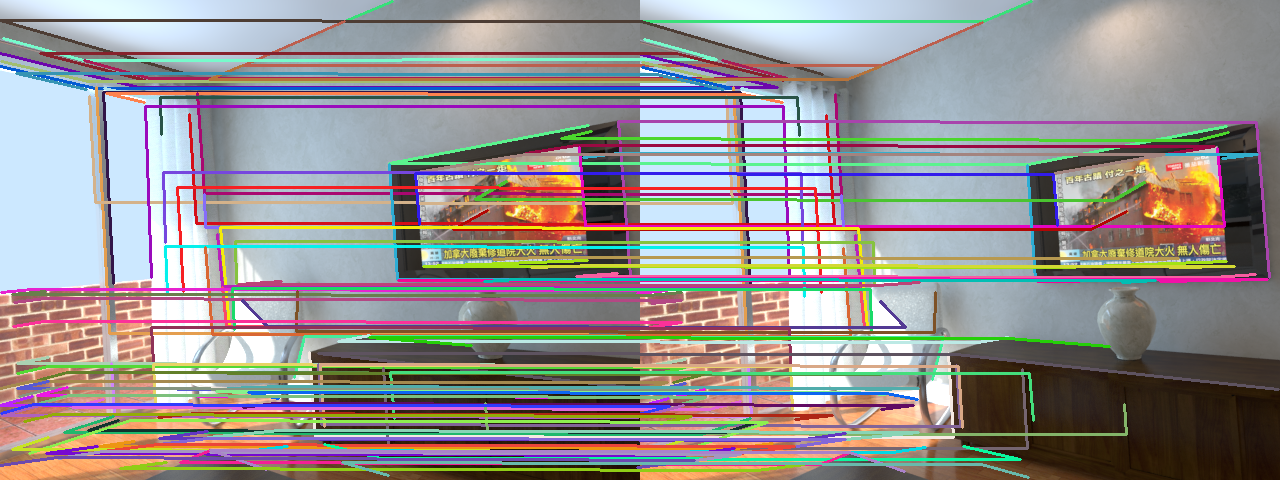}
	\includegraphics[width=0.150\textheight]{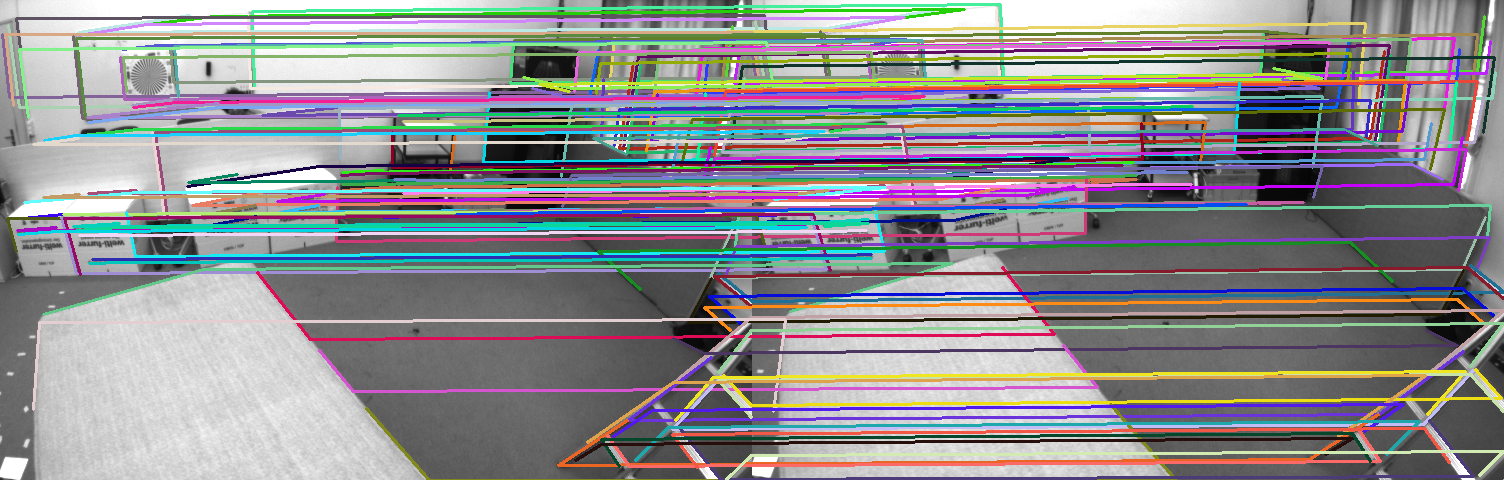}
	\includegraphics[width=0.315\textheight]{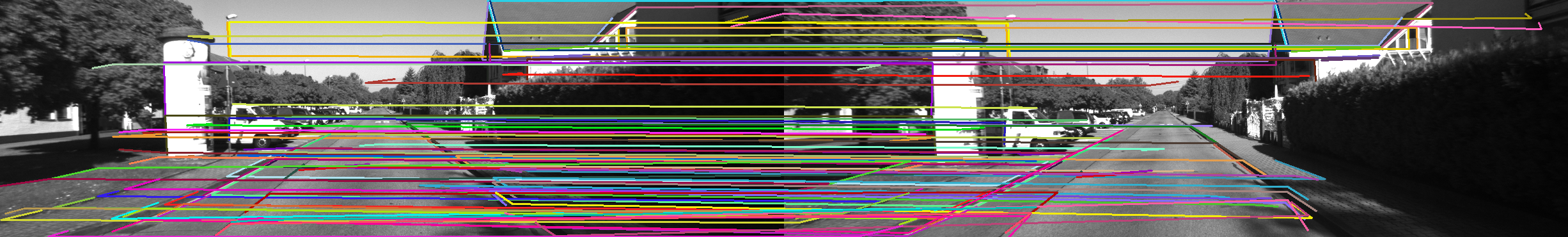}}	
	
\caption{Examples of line feature tracking of LSD+LBD+KNN and LSD+TET (ours). From left to right, image pairs are from TUM\cite{tum}, ICL-NUIM \cite{icl}, Euroc \cite{euroc}, and KITTI \cite{kitti} with, respectively. Intuitively, our method tracks more matching pairs than LSD+LBD+KNN.} \label{f:trackmatch}
\end{figure*}

\begin{figure}[tp]
\centering  
	\includegraphics[width=0.4\textwidth]{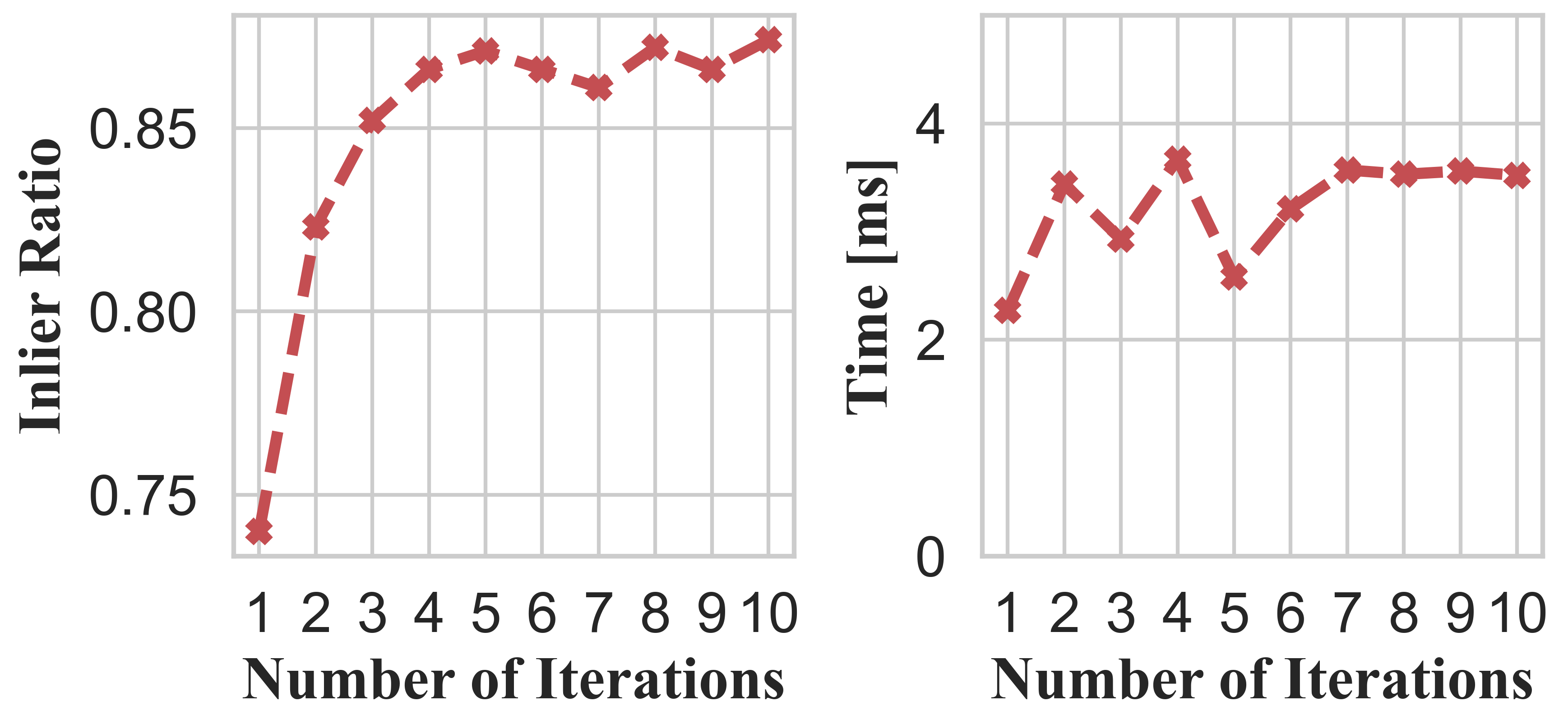}
	\quad
\caption{The comparsion of inlier ratio and tracking time on the EuRoc/V1-02 sequence. In this case, the residual function usually takes about 3.7 ms to converge after 5 iterations. In consideration of generality, this work sets $N_{itermax} = 10$ .
}\label{f:iteratation}
\end{figure}

\subsection{Image Pyramid}
To improve robustness, a $n$-level image pyramid scheme is adopted in this work, which allows algorithm to estimate a small movement vector at the deepest layer at first. Let $j\in {1,2,...,n}$, $I_{1}^{j}$ and $I_{2}^{j}$ denote two corresponding images at the $j$-th layer, and scale ratio $=2$ between two adjacent layers. Take $I_{1}$ as an example, the images in the pyramid can be described as:
\begin{equation}
\begin{aligned}
& I_{1}^{j}=  \frac{1}{4}I_{1}^{j-1}(2x,2y)+ 
\\& \frac{1}{8}\left((I_{1}^{j-1}(2x\pm1,2y)+I_{1}^{j-1}(2x,2y\pm1)\right)+
\\& \frac{1}{16}\left((I_{1}^{j-1}(2x\pm1,2y\pm1)\right)
\end{aligned},
\label{e:pyramid}
\end{equation}
where $\pm$ includes two cases $+$ and $-$, e.g., $(2x\pm1,2y)=(2x+1,2y)+(2x-1,2y)$. And then let $L_{1}^{j}$ denote corresponding line features, the coordinate of its two endpoints are obtained by:
\begin{equation}
\begin{cases}
L_{1}^{j}(x_{1}^{},y_{1}^{})=\frac{L_{1}(x_{1}^{},y_{1}^{})}{2^j} 
\\ L_{1}^{j}(x_{2}^{},y_{2}^{})=\frac{L_{1}(x_{2}^{},y_{2}^{})}{2^j}.
\end{cases}
\label{e:linepy}
\end{equation}\par
The movement vectors are estimated in the image pyramid, which is described as the third step in \textbf{Algorithm} 1. In this work, we set $n = 3$.

\section{Experiments}  
In this section, we evaluate the efficency of the proprosed LSD+TET method on four public benchmark datasets including ICL-NUIM \cite{icl}, TUM\cite{tum}, Euroc \cite{euroc}, and KITTI \cite{kitti} with resolution 640$\times$480, 640$\times$480, 752$\times$480, and 1242$\times$375, respectively. Our TET method was implemented using C++ based on OpenCV. All experiments were performed on the Intel Core CPU i7-10710U @1.10 GHz.

\begin{table}[]
\caption{Tracking performance of popular LSD+LBD+KNN method and our LSD+TET method.}
\label{t:track}
\setlength{\tabcolsep}{1.1mm}{
\begin{tabular}{@{}lcccccc@{}}
\toprule
\multicolumn{1}{c}{\multirow{2}{*}{Dataset}} & \multicolumn{3}{c}{LSD+LBD+KNN} & \multicolumn{3}{c}{LSD+TET}    \\ \cmidrule(l){2-7} 
\multicolumn{1}{c}{}                         & Time     & Number    & Ratio    & Time        & Number & Ratio   \\ \midrule
TUM/freiburg1-desk                           & 53       & 27        & 80\%     & 8           & 84     & 78\%    \\
TUM/freiburg1-xyz                            & 49       & 24        & 87\%     & 9           & 77     & 93\%    \\
ICL-NUIM/office1                             & 45       & 34        & 82\%     & 15          & 36     & 80\%    \\
ICL-NUIM/livingroom1                         & 51       & 72        & 92\%     & 8           & 82     & 95\%    \\
EuRoc/V1-02                                  & 53       & 122       & 82\%     & 14          & 95     & 89\%    \\
EuRoc/MH-05                                  & 54       & 115       & 87\%     & 13          & 86     & 76\%    \\
KITTI/06                                     & 102      & 21        & 83\%     & 20          & 48     & 84\%    \\
KITTI/07                                     & 103      & 27        & 87\%     & 21          & 53     & 79\%    \\
Average                                      & 63.75    & 55.25     & 85\%     & \textbf{13.5} & 70.125 & 84.25\% \\ \bottomrule
\end{tabular}
}
\begin{flushleft}
Time denotes the average tracking time per frame. Number denotes the number of the matching line feature pairs after descriptor distance (LSD+LBD+KNN) or termination condition (LSD+TET) culling. Ratio means the inlier ratio after RANSAC. Our method achieves highly competitive accuracy with dominant advantage over tracking time. 
\end{flushleft}
\end{table}

\subsection{Line Feature Tracking}

In this subsection we test the monocular tracking performance of the proposed LSD+TET method by comparing it with the SOTA LSD+LBD+KNN method. Note that we set the condition of re-detecting line feature as if current inliers is less than 30. We observe that our method only needs to detect once in five frames as the descriptor-based methods need to detect per frame. \par
Table \ref{t:track} provides a comparison of tracking performance of LSD+LBD+KNN and our LSD+TET method, we can conclude that:
\begin{itemize}
	\item From the last row, generally, our LSD+TET LSD+TET method takes 13.5 ms per frame on average while LSD+LBD+KNN 63.75 ms. Therefore, our method	has a prominent advantage over time consumption and it is nearly 5 times as fast as LSD+LBD+KNN. 
	\item Besides, our method achieves competitive accuracy in terms of the inliers ratio evaluation.	
\end{itemize}\par
However, it needs to point out that our method produces a relatively low accuracy on TUM/freiburg1-desk and EuRoc/MH-05, the former represents the fast-motion sequence and the latter represents the low-light sequence. 
\par
In addition, Fig \ref{f:trackmatch} shows same example of line feature tracking between two adjacent frames on the all four datasets. The images from KITTI have the highest resolution, so, they usually need relatively high computation to detect and track line features.

\begin{figure}[tp]
\centering  
	\includegraphics[width=0.23\textwidth]{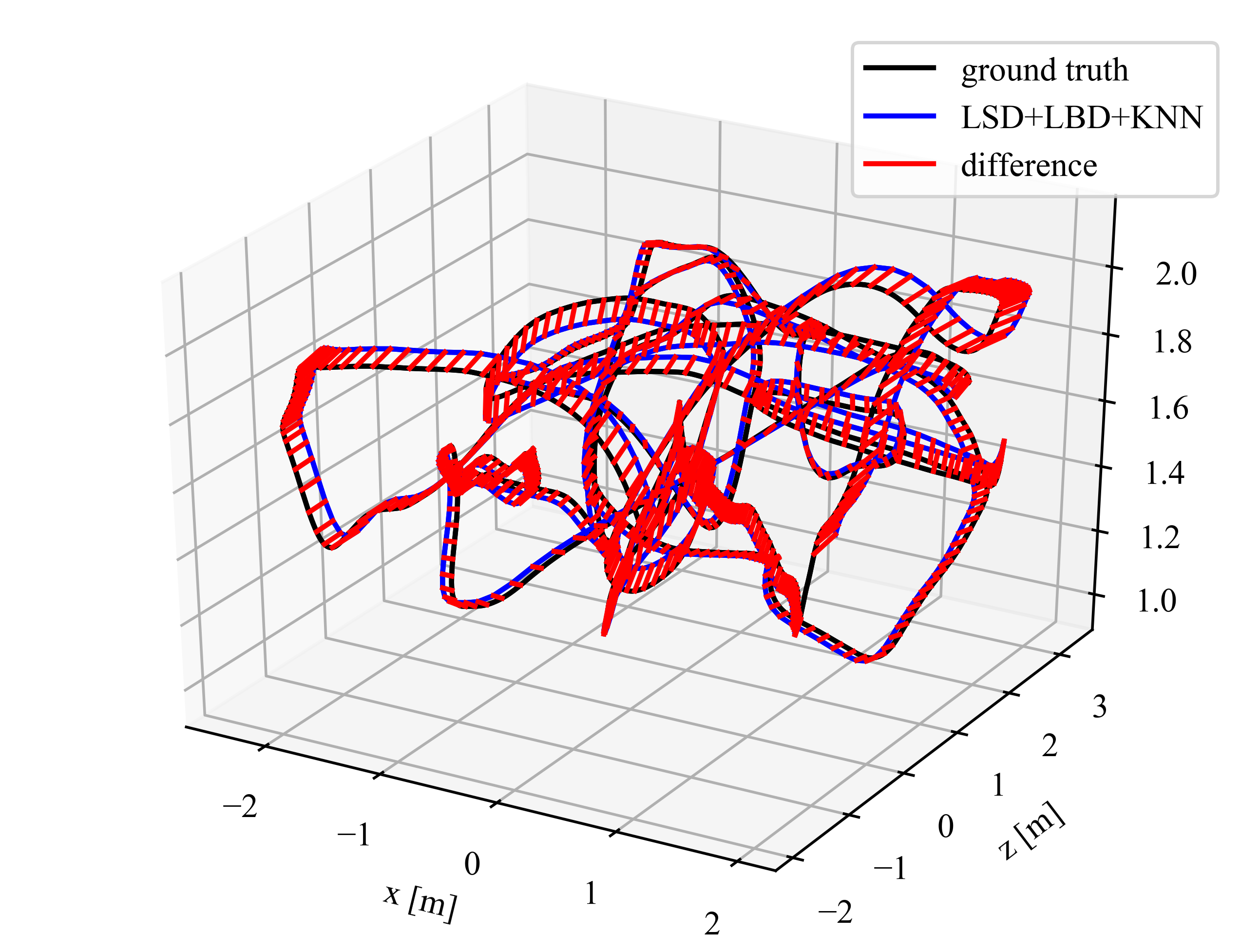}
	\includegraphics[width=0.23\textwidth]{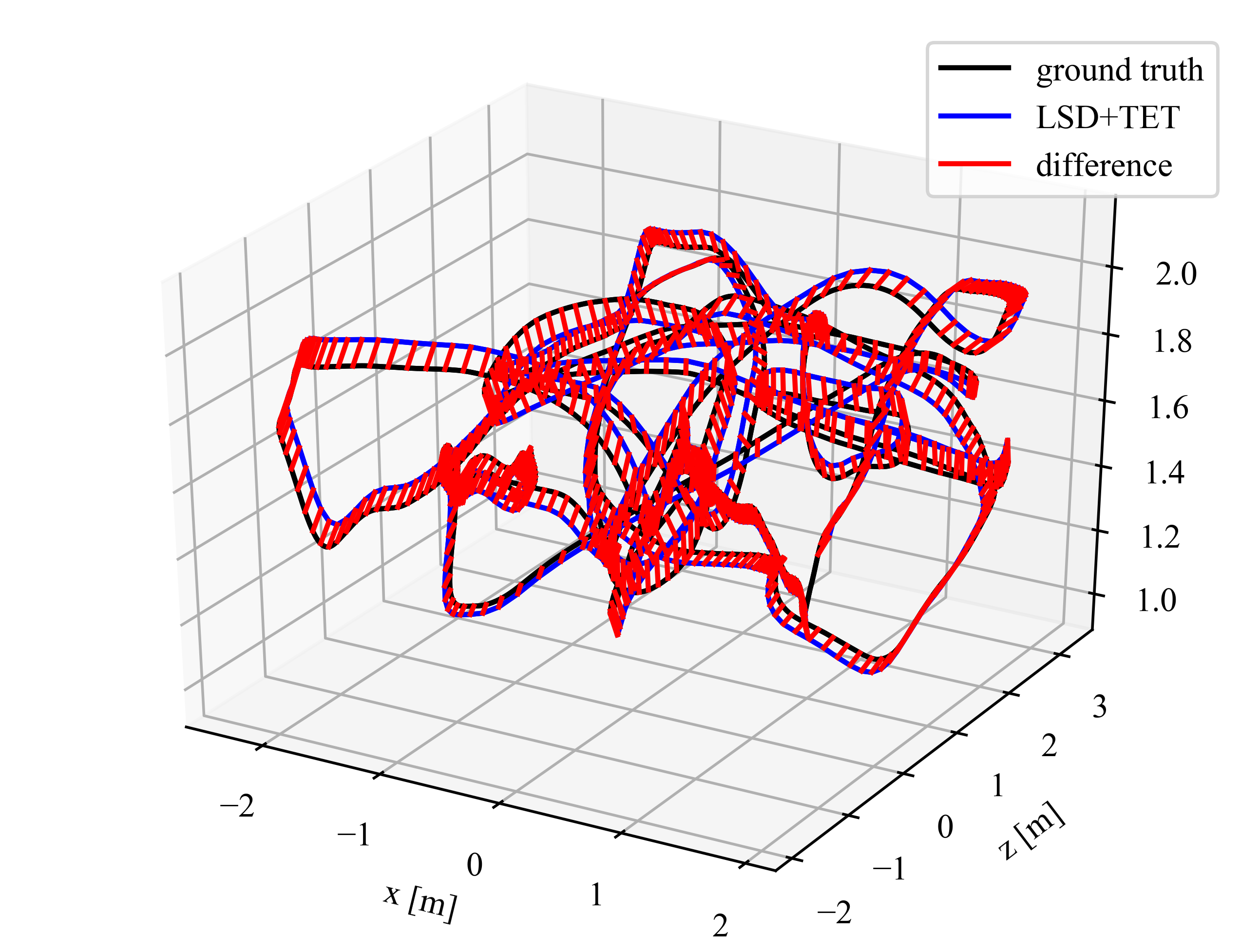}
\quad
\caption{Trajectory comparison of the open-source PL-VINS method 
using LSD+LBD+KNN or LSD+TET on the EuRoc/V1-02 sequence. In this case, LSD+TET yields highly competitive accuracy in terms of the APE evaluation on this sequence.}
\label{linevector}
\end{figure}

\subsection{Application in SLAM}
In this subsection, we make an attempt to apply LSD+TET for the SLAM problem. To be specific, the attempt is implemented by leveraging the LSD+TET method into an open-source SLAM solution\footnote{https://github.com/cnqiangfu/PL-VINS} that incorporates line features, and we refer the SLAM solution as PL-VINS. \par 
Fig \ref{f:methodcomparsion} provides a comparison of three-dimension motion trajectory. Fruther, we use the RMSE value of absolute trajectory error (APE) \cite{tum} to quantize the result. In which, LSD+TET yields 0.0927 m while LSD+LBD+KNN 0.102 m. We can conclude our method produces highly competitive accuracy in terms of the APE evaluation. However, it also needs to point out that our method has an obvious advantage over tracking time consumption.

\section{Conclusions}
In this paper, we presents a novel light-weight sparse optical flow-based line feature tracking method. By re-defining the tracking goal as to track two endpoints of a line feature, an efficient two endpoint tracking (TET) method is proposed to achieve this goal. We test our method by comparing it to the SOTA LSD+LBD+KNN method on the indoor scenes including six sequences from TUM, ICL-NUIM, and EuRoc dataset, and the outdoor scenes including two sequences from KITTI dataset. The result shows our method achieves competitive line feature tracking performance in terms of inlier ratio with nearly 5 time tracking speed. Besides, we also make a simple application attempt by implementing LSD+TET to an open-source SLAM method, which also produces SOTA ATE accuracy on the EuRoc/V1-02 sequence. Our method is really novel as it is obviously different from current descriptor-based methods, which has a dominant advantage over computation as it needs not to compute descriptors or detect line feature repeatedly.\par
It needs to point out that our method is not able to handle with the fast motion and low light situation well. For future works, we plan to improve it for the two situations by exploiting the motion prediction and auto exposure technologies.   



\begin{thebibliography}{100}
\bibitem{poseline}Abdellali, Hichem, Robert Frohlich, and Zoltan Kato. "Robust Absolute and Relative Pose Estimation of a Central Camera System from 2D-3D Line Correspondences." Proceedings of the IEEE International Conference on Computer Vision Workshops. 2019.

\bibitem{pnl}Lecrosnier, Louis, et al. "Camera pose estimation based on PnL with a known vertical direction." IEEE Robotics and Automation Letters 4.4 (2019): 3852-3859.

\bibitem{Li}Li, Haoang, et al. "Line-based absolute and relative camera pose estimation in structured environments." 2019 IEEE/RSJ International Conference on Intelligent Robots and Systems (IROS). IEEE, 2019.

\bibitem{lpcvpr}Fabbri, Ricardo, et al. "TRPLP-Trifocal Relative Pose From Lines at Points." Proceedings of the IEEE/CVF Conference on Computer Vision and Pattern Recognition. 2020.

\bibitem{xupami}Xu, Chi, et al. "Pose estimation from line correspondences: A complete analysis and a series of solutions." IEEE transactions on pattern analysis and machine intelligence 39.6 (2016): 1209-1222.

\bibitem{fu2}Fu, Qiang, et al. "A robust RGB-D SLAM system with points and lines for low texture indoor environments." IEEE Sensors Journal 19.21 (2019): 9908-9920.

\bibitem{stuctvio}Zou, Danping, et al. "StructVIO: visual-inertial odometry with structural regularity of man-made environments." IEEE Transactions on Robotics 35.4 (2019): 999-1013.

\bibitem{plvio}He, Yijia, et al. "Pl-vio: Tightly-coupled monocular visual-inertial odometry using point and line features." Sensors 18.4 (2018): 1159.

\bibitem{yangIROS}Yang, Yulin, et al. "Visual-Inertial Odometry with Point and Line Features." 2019 IEEE/RSJ International Conference on Intelligent Robots and Systems (IROS). IEEE, 2019.

\bibitem{yangtro}Yang, Yulin, and Guoquan Huang. "Observability analysis of aided ins with heterogeneous features of points, lines, and planes." IEEE Transactions on Robotics 35.6 (2019): 1399-1418.

\bibitem{plslam}Gomez-Ojeda, Ruben, et al. "PL-SLAM: A stereo SLAM system through the combination of points and line features." IEEE Transactions on Robotics 35.3 (2019): 734-746.

\bibitem{strifovio}Zheng, Feng, et al. "Trifo-VIO: Robust and efficient stereo visual inertial odometry using points and lines." 2018 IEEE/RSJ International Conference on Intelligent Robots and Systems (IROS). IEEE, 2018.

\bibitem{plvins}Fu, Qiang, et al. "PL-VINS: Real-Time Monocular Visual-Inertial SLAM with Point and Line." arXiv preprint arXiv:2009.07462 (2020).

\bibitem{zhang}Zhang, Guoxuan, et al. "Building a 3-D line-based map using stereo SLAM." IEEE Transactions on Robotics 31.6 (2015): 1364-1377.

\bibitem{fastorbslam} Q. Fu et al., "Fast ORB-SLAM Without Keypoint Descriptors," in IEEE Transactions on Image Processing, vol. 31, pp. 1433-1446, 2022, doi: 10.1109/TIP.2021.3136710.

\bibitem{vins}Qin, Tong, Peiliang Li, and Shaojie Shen. "Vins-mono: A robust and versatile monocular visual-inertial state estimator." IEEE Transactions on Robotics 34.4 (2018): 1004-1020.

\bibitem{fu1}Yu, Hongshan, et al. "Robust robot pose estimation for challenging scenes with an RGB-D camera." IEEE Sensors Journal 19.6 (2018): 2217-2229.

\bibitem{cnnsvo}Loo, Shing Yan, et al. "CNN-SVO: Improving the mapping in semi-direct visual odometry using single-image depth prediction." 2019 International Conference on Robotics and Automation (ICRA). IEEE, 2019.

\bibitem{linemgeo}Gomez-Ojeda, Ruben, and Javier Gonzalez-Jimenez. "Geometric-based line feature tracking for HDR stereo sequences." 2018 IEEE/RSJ International Conference on Intelligent Robots and Systems (IROS). IEEE, 2018.

\bibitem{lsd}Von Gioi, Rafael Grompone, et al. "LSD: A fast line segment detector with a false detection control." IEEE transactions on pattern analysis and machine intelligence 32.4 (2008): 722-732.

\bibitem{lbd}Zhang, Lilian, and Reinhard Koch. "An efficient and robust line feature matching approach based on LBD descriptor and pairwise geometric consistency." Journal of Visual Communication and Image Representation 24.7 (2013): 794-805.

\bibitem{knn}Steinbach, Michael, and Pang-Ning Tan. "kNN: k-nearest neighbors." The top ten algorithms in data mining (2009): 151-162.


\bibitem{klt}Baker, Simon, and Iain Matthews. "Lucas-kanade 20 years on: A unifying framework." International journal of computer vision 56.3 (2004): 221-255.

\bibitem{icl}Handa, Ankur, et al. "A benchmark for RGB-D visual odometry, 3D reconstruction and SLAM." 2014 IEEE international conference on Robotics and automation (ICRA). IEEE, 2014.

\bibitem{detect1}Liu, Yang, Zongwu Xie, and Hong Liu. "LB-LSD: A length-based line segment detector for real-time applications." Pattern Recognition Letters 128 (2019): 247-254.

\bibitem{detect2}Zhang, Ziheng, et al. "Ppgnet: Learning point-pair graph for line segment detection." Proceedings of the IEEE Conference on Computer Vision and Pattern Recognition. 2019.

\bibitem{detect3}Xue, Nan, et al. "Learning attraction field representation for robust line segment detection." Proceedings of the IEEE Conference on Computer Vision and Pattern Recognition. 2019.

\bibitem{detect4}Huang, Siyu, et al. "TP-LSD: Tri-Points Based Line Segment Detector." arXiv preprint arXiv:2009.05505 (2020).

\bibitem{tracking1}Wei, Dong, Yongjun Zhang, and Chang Li. "Robust Line Segment Matching via Reweighted Random Walks on the Homography Graph." Pattern Recognition (2020): 107693.

\bibitem{tracking2}Vakhitov, Alexander, and Victor Lempitsky. "Learnable line segment descriptor for visual SLAM." IEEE Access 7 (2019): 39923-39934.


\bibitem{mvg}Hartley, Richard, and Andrew Zisserman. Multiple view geometry in computer vision. Cambridge university press, 2003.

\bibitem{tum}Sturm, Jürgen, et al. "A benchmark for the evaluation of RGB-D SLAM systems." 2012 IEEE/RSJ International Conference on Intelligent Robots and Systems. IEEE, 2012.

\bibitem{euroc}Burri, Michael, et al. "The EuRoC micro aerial vehicle datasets." The International Journal of Robotics Research 35.10 (2016): 1157-1163.

\bibitem{kitti}Geiger, Andreas, et al. "Vision meets robotics: The kitti dataset." The International Journal of Robotics Research 32.11 (2013): 1231-1237.

\end{thebibliography}
\end{document}